\pdfoutput=1

\documentclass[11pt]{article}

\usepackage[final]{acl}

\usepackage{times}
\usepackage{latexsym}

\usepackage[T1]{fontenc}

\usepackage[utf8]{inputenc}

\usepackage{microtype}

\usepackage{inconsolata}

\usepackage{graphicx}

\usepackage{threeparttable}
\usepackage{booktabs}
\usepackage{multirow}
\usepackage{multicol}
\usepackage{amsmath}
\usepackage{tabularx}
\usepackage{colortbl}
\usepackage{subcaption}

\title{SEOE: A Scalable and Reliable Semantic Evaluation Framework for Open Domain Event Detection}

\author{
 \textbf{Yi-Fan Lu},
 \textbf{Xian-Ling Mao},
 \textbf{Tian Lan},
 \textbf{Tong Zhang},
 \textbf{Yu-Shi Zhu},
 \textbf{Heyan Huang}
\\
 Beijing Institute of Technology
\\
\texttt{\href{mailto:yifanlu@bit.edu.cn}{yifanlu@bit.edu.cn}},
\texttt{\href{mailto:maoxl@bit.edu.cn}{maoxl@bit.edu.cn}},
\texttt{\href{mailto:lantiangmftby@gmail.com}{lantiangmftby@gmail.com}},
\\
\texttt{\href{mailto:tongz@bit.edu.cn}{tongz@bit.edu.cn}},
\texttt{\href{mailto:zhuyushi@bit.edu.cn}{zhuyushi@bit.edu.cn}},
\texttt{\href{mailto:hhy63@bit.edu.cn}{hhy63@bit.edu.cn}}
\\
\texttt{https://github.com/Lyfralston/SEOE}
}

\begin{document}
\maketitle
\begin{abstract}
Automatic evaluation for Open Domain Event Detection (ODED) is a highly challenging task, because ODED is characterized by a vast diversity of un-constrained output labels from various domains. 
Nearly all existing evaluation methods for ODED usually first construct evaluation benchmarks with limited labels and domain coverage, and then evaluate ODED methods using metrics based on token-level label matching rules. 
However, this kind of evaluation framework faces two issues: (1) The limited evaluation benchmarks lack representatives of the real world, making it difficult to accurately reflect the performance of various ODED methods in real-world scenarios; (2) Evaluation metrics based on token-level matching rules fail to capture semantic similarity between predictions and golden labels.
To address these two problems above, we propose a scalable and reliable \textbf{S}emantic-level \textbf{E}valuation framework for \textbf{O}pen domain \textbf{E}vent detection (SEOE) by constructing a more representative evaluation benchmark and introducing a semantic evaluation metric. 
Specifically, our proposed framework first constructs a scalable evaluation benchmark that currently includes 564 event types covering 7 major domains, with a cost-effective supplementary annotation strategy to ensure the benchmark's representativeness. The strategy also allows for the supplement of new event types and domains in the future. 
Then, the proposed SEOE leverages large language models as automatic evaluation agents to compute a semantic F1-score, incorporating fine-grained definitions of semantically similar labels to enhance the reliability of the evaluation. 
Extensive experiments validate the representatives of the benchmark and the reliability of the semantic evaluation metric. Existing ODED methods are thoroughly evaluated, and the error patterns of predictions are analyzed, revealing several insightful findings. Our benchmark and evaluation toolkits are publicly available.
\end{abstract}

\section{Introduction}

\begin{figure}[ht]
    \centering
    \includegraphics[width=\linewidth]{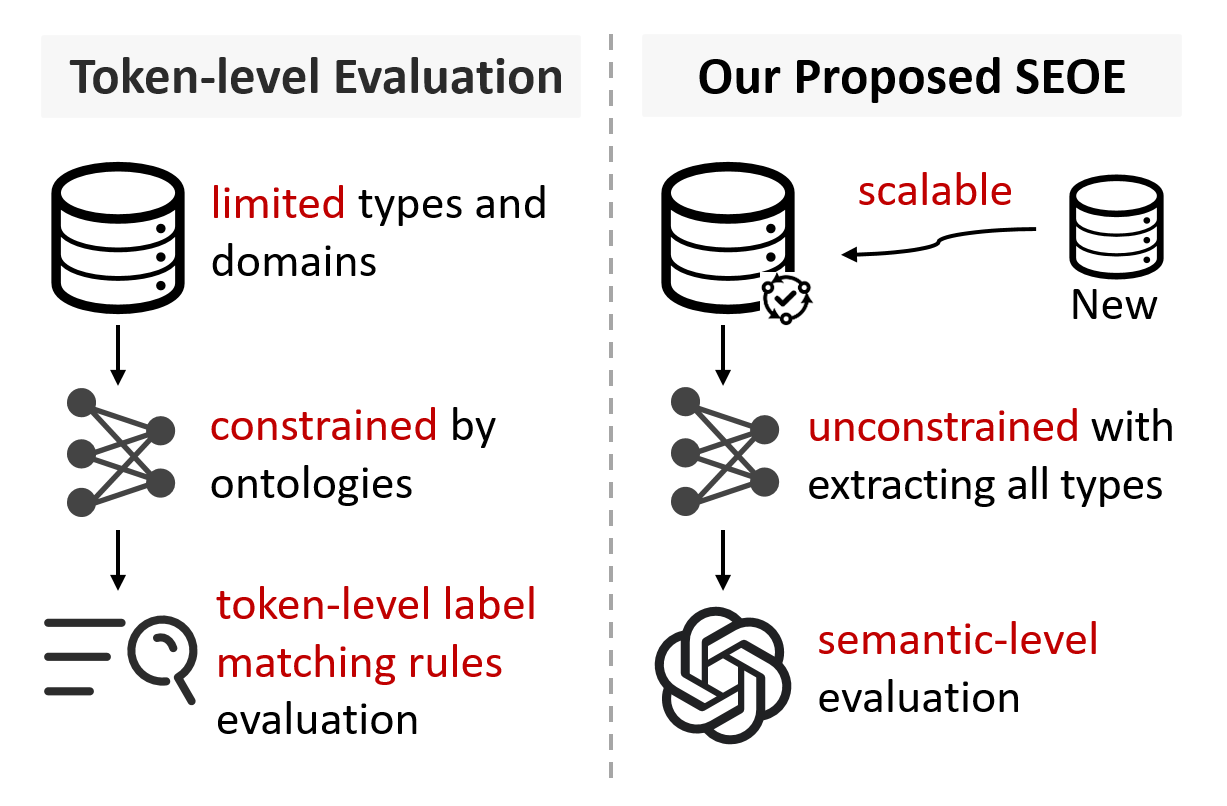}
    \caption{Comparison between previous ODED evaluation frameworks and our proposed SEOE.}
    \label{fig:intro}
\end{figure}

Open domain event detection (ODED) is a crucial task in open domain information extraction.
Its primary objective is to extract structured information, namely events, from large-scale textual data across various domains without constrained ontologies.
This structured information can then be used for further research tasks, such as information retrieval \cite{zhang-etal-2024-event, joshi-etal-2023-u}, 
knowledge base construction \cite{10.1145/3654522.3654533}, and large language models \cite{cai-etal-2024-improving-event, Chen_Qin_Jiang_Choi_2024, wang2023instructuiemultitaskinstructiontuning}, where understanding the relationships between events is essential. 
Recently, the importance of ODED has increased with the rapid expansion of the web, which continuously generates vast amounts of unstructured information across various domains, such as the clinical area \cite{ma-etal-2023-dice}, suicide understanding \cite{guzman-nateras-etal-2022-event}, and scientific publications \cite{lu2024exceedsextractingcomplexevents}. 

Evaluating ODED is a highly challenging task due to the vast diversity of event types in the open domain \cite{araki-mitamura-2018-open}. When making predictions, ODED models are often required to extract events with unseen event types during the training and validation stage \cite{cai-etal-2024-improving-event}.
To evaluate these models, to the best of our knowledge, nearly all existing ODED evaluation frameworks simulate the open setting by evaluating models using a predefined ontology in a zero-shot manner \cite{cai-etal-2024-improving-event, yue-etal-2023-zero}. 
Specifically, as illustrated in Figure~\ref{fig:intro}, these frameworks typically first construct or utilize evaluation benchmarks with a limited set of types and domain coverage that are unseen during the training stage.
During the evaluation process, ODED models are required to recognize events constrained by predefined ontologies of these benchmarks. These evaluation frameworks then apply token-level matching rules to evaluate models' predictions, such as strict match \cite{zhang-etal-2022-efficient-zero}, partial match \cite{araki-mitamura-2018-open}, and head noun match \cite{li-etal-2021-document}.

However, the above ODED evaluation frameworks face two issues: 
(1) The evaluation benchmarks, with limited labels and domain coverage, lack representatives of the real world. Considering such benchmarks as the simulation of the open domain, it is difficult to accurately reflect the performance of ODED models in real-world scenarios; 
(2) Evaluation metrics based on token-level matching rules fail to capture semantic similarity between predictions and golden event types. Numerous semantically similar event types in open domain cannot be accurately evaluated under these evaluation rules.

To address these two problems above, we propose a scalable and reliable \textbf{S}emantic \textbf{E}valuation framework for \textbf{O}pen domain \textbf{E}vent detection (SEOE). This framework consists of a more representative evaluation benchmark and a semantic evaluation metric. 
As for the benchmark, our proposed framework SEOE constructs a scalable evaluation benchmark that currently includes 564 event types covering 7 major domains. 
We first integrate the ontologies of mainstream event extraction datasets into a larger ontology. Then, we propose a cost-effective supplementary annotation strategy to ensure that each data instance is examined by the integrated ontology, making the benchmark both representative of real-world scenarios and easily scalable.
As for the evaluation metric, SEOE leverages large language models (LLMs) as automatic evaluation agents to compute a semantic F1-score. 
To ensure reliable evaluation, SEOE generates fine-grained definitions of event types and groups semantically similar labels together to assist in the evaluation process.

As for the experiments, the representatives of the evaluation benchmark and the reliability of the semantic evaluation metric are validated. 
With such a representative benchmark and a reliable metric in SEOE, we conduct comprehensive evaluation experiments on existing ODED methods including closed-source and open-source LLMs. 
Extensive experiments reveal that ODED remains a highly challenging task, particularly in balancing the accuracy and diversity of the predictions.
Furthermore, a systematic analysis about the patterns of the model's incorrect predictions reveals that predicting event types and generating their definitions in the ODED task is a highly challenging problem.

In summary, the contributions of our study are two-fold:

\begin{itemize}
    \item \textbf{SEOE}: We propose a scalable and reliable semantic evaluation framework for ODED, which is built upon a more representative evaluation benchmark and a semantic evaluation metric, with LLM serving as the automatic evaluation agent. SEOE achieves cost-effective scalability and demonstrates strong correlations with human evaluations.
    \item \textbf{Comprehensive evaluation and analysis}: We apply our proposed SEOE to comprehensively evaluate the performance of existing ODED models. We systematically analyze the patterns of the models' incorrect predictions. The experiments reveal insightful phenomenon worth further investigation.
\end{itemize}

\section{Related Works}

\subsection{Event Detection}
Event detection (ED), as an important sub-task of event extraction in information extraction, can be divided into closed-domain ED and open-domain ED based on the presence of a predefined event schema \cite{liu2021overvieweventextractionapplications}. 
Due to the increasing number of domains of interest \cite{sun-etal-2022-phee} and expensive expert annotations when constructing closed-domain datasets \cite{hsu-etal-2022-degree}, ODED research has gained significant attention \cite{veyseh2021augmenting, liu-etal-2019-open}. 
Besides, some studies on few-shot \cite{ma-etal-2023-shot} and zero-shot ED \cite{yue-etal-2023-zero, zhang-etal-2022-zero} can emulate the open-domain setting. 
However, these studies provide predefined event types during model prediction, which does not align with real-world open-domain scenarios.

    

\subsection{LLMs for Automatic Benchmark Construction}
With the emergence of advanced LLMs, exemplified by GPT-4, automated data annotation and synthesis have garnered significant attention in the NLP community \cite{tan-etal-2024-large}. Many studies on LLM-based annotation focus on automating the labeling process with the help of LLMs \cite{yadav2024automatingtextannotationcase, tseng2024expertlevellanguagemodelsexpertlevel}, including single LLM annotation \cite{Chen_Qin_Jiang_Choi_2024, 10938896}, joint annotation by multiple LLMs \cite{martorana2024zero}, and hybrid annotation by humans and LLMs \cite{li-etal-2023-coannotating}. In addition, the evaluation of annotations is also a crucial part of automated annotation, with mainstream studies including general evaluation with human comparisons \cite{honovich-etal-2023-unnatural}, 
task-specific evaluations  \cite{chen-etal-2023-disco}, 
and using LLMs as evaluators known as LLM-as-a-judge \cite{wu2024metarewardinglanguagemodelsselfimproving, zheng2023judging}. 
In this paper, we propose a cost-effective method to instruct GPT-4o to construct a representative ODED benchmark. 

\subsection{Evaluation for Event Detection}
ED is typically evaluated using token-level matching rules~\cite{li-etal-2021-document, lin-etal-2020-joint}, which assess whether the model-generated events and their labels are token-level identical or similar to the human-annotated ground truth. However, these rules exhibit weak correlation with human judgments due to the lack of the semantic evaluation \cite{lu2024exactmatchsemanticallyreassessing}.
Recently, there is an emerging trend in the natural language generation research community to leverage LLMs for evaluating generations in a semantic manner~\cite{li2025generationjudgmentopportunitieschallenges, lan2024traininglanguagemodelscritique,lan2020pone}. Notably, \citep{lu2024exactmatchsemanticallyreassessing} propose a LLM-based automatic evaluation framework for event extraction, named RAEE. 
However, RAEE is designed for closed-domain scenarios and faces the challenge of evaluating a wide range of event types with varying definitions in open-domain scenarios.
In this paper, we propose a LLM-based automatic evaluation framework that includes a scalable benchmark and a reliable evaluation metric to semantically evaluate ODED models.

\section{ODED Task Formulation}
\label{sec:task_formulation}
To the best of our knowledge, nearly all existing ODED evaluation frameworks leverage limited evaluation benchmarks to evaluate ODED models with constrained event types in a zero-shot way. This kind of evaluation framework cannot reflect ODED methods' performance in real-world scenarios. Therefore, it is necessary to formally define the ODED task for further evaluation.

Given a text sequence $T = \{ w_1, w_2 , \cdots, w_k\}$, where $w_i$ represents $i$-th word in the text, ODED models are tasked with identifying event triggers $\mathcal{E}=\{e_1, e_2, \cdots, e_m\}$. 
Each event trigger mention comprises words from $T$. 
In addition to identifying event trigger mentions, models are required to predict event types, denoted as $\mathcal{C}$, for each trigger, without relying on pre-defined ontologies.
Furthermore, models must provide fine-grained definitions $\mathcal{D} = \{(t,d) \mid t \in \mathcal{C}\}$ for each event type.


\section{Open Domain Event Benchmark}
\label{sec:benchmark}
Nearly all existing ODED evaluation benchmarks are limited in the diversity of event types and domains, which are not representative of real-world scenarios. To solve this problem, we propose a scalable and more representative benchmark.
We will first introduce the construction process in Section~\ref{sec:benchmark_construction} and then analyze it in Section~\ref{sec:benchmark_reliability} and \ref{sec:benchmark_analysis}.

\subsection{Benchmark Construction}
\label{sec:benchmark_construction}

\begin{figure*}[ht]
    \centering
    \includegraphics[width=\linewidth]{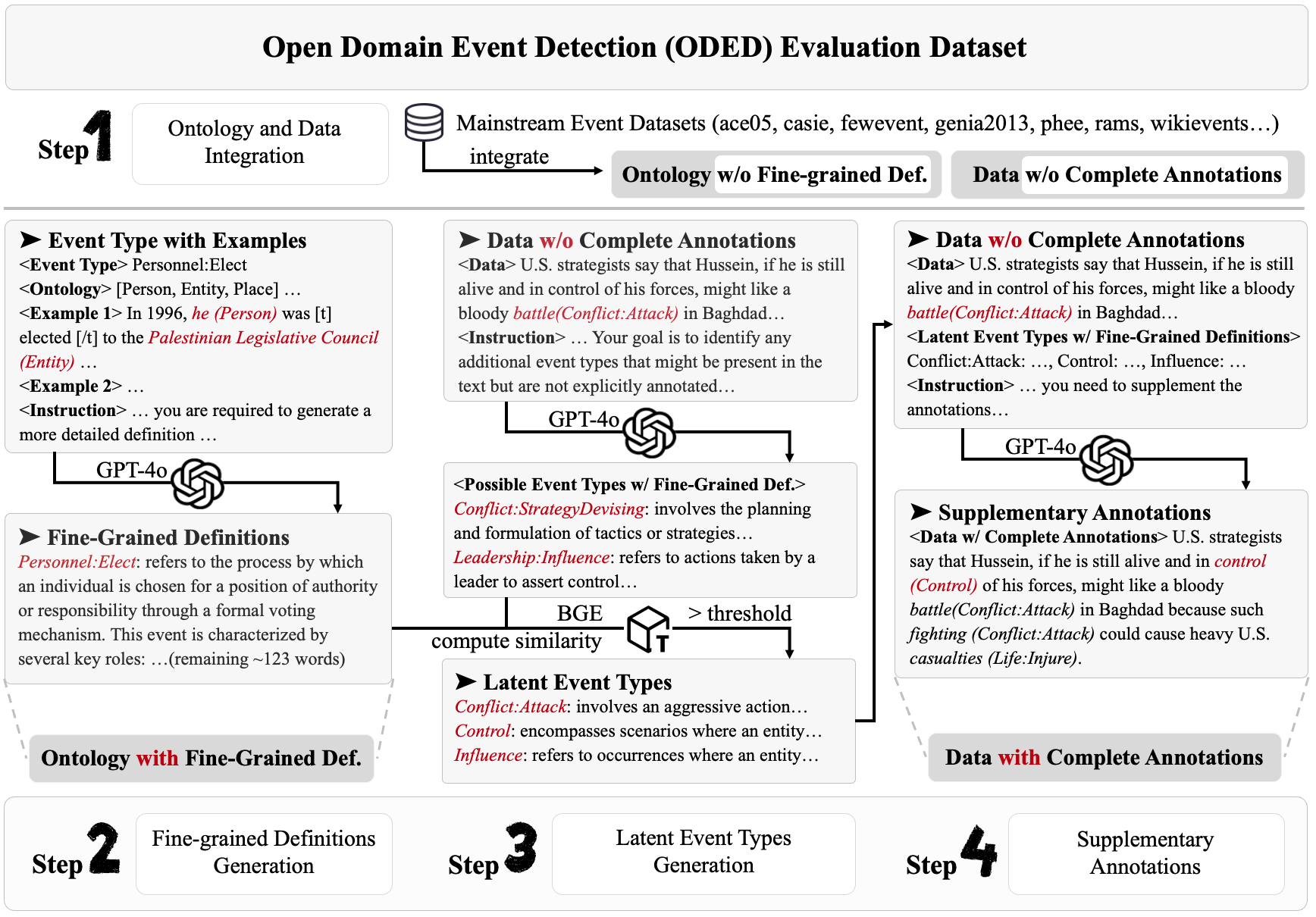}
    \caption{The pipelined construction process of proposed open domain event detection evaluation dataset. }
    \label{fig:benchmark_construction}
\end{figure*}

For a comprehensive benchmark, it is widely agreed that each data instance should be examined across all predefined types, regardless of whether these types are ultimately annotated \cite{zhang2013review}.  
However, this requirement is extremely costly for an ODEE evaluation benchmark, given the large number of event types across various domains and the potential scalability of the benchmark in terms of event types.
To solve this issue, we leverage GPT-4o\footnote{https://openai.com/index/hello-gpt-4o/} to enable automatic annotation and propose a cost-effective method to ensure that each data instance is examined by all event types from an integrated ontology. 
Figure~\ref{fig:benchmark_construction} illustrates the process of benchmark construction in the following four steps.
Related prompts can be found in Appendix~\ref{sec:benchmark_construction_prompt}.

\begin{figure}
    \centering
    \includegraphics[width=\linewidth]{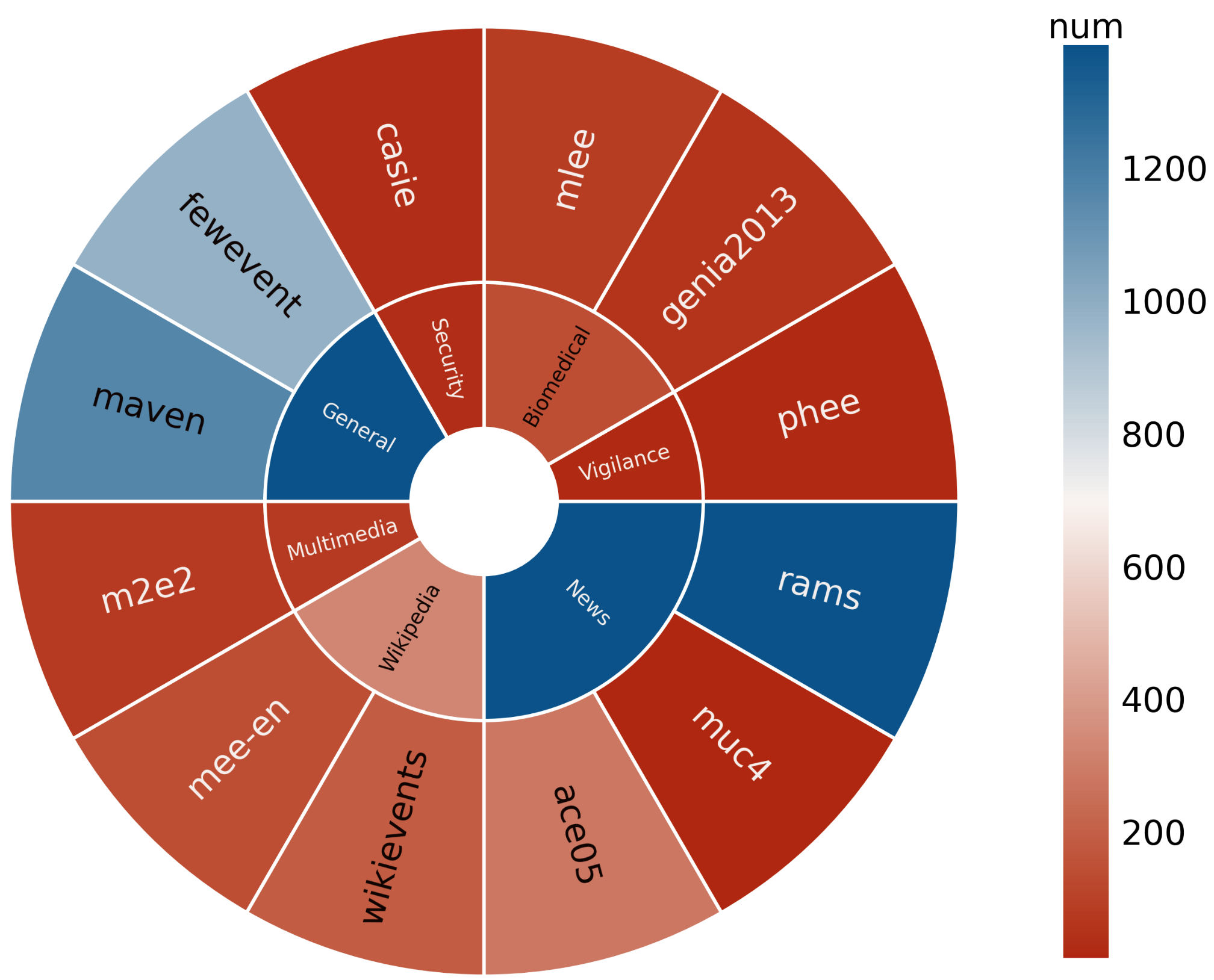}
    \caption{Distribution of SEOE benchmark.}
    \label{fig:data_distribution}
\end{figure}

\paragraph{Step 1: Ontology and Data Integration}
It is important for an ODED dataset to include a diversity of event types and a breadth of domains. Similar to UniversalNER \cite{zhou2023universalner}, we integrate mainstream event datasets' ontologies into a large ontology and sample a subset of the data by uniformly sampling cases across all types.
More details about the integration is shown in Appendix~\ref{sec:benchmark_integration}.
An overview of data distribution is shown in Figure~\ref{fig:data_distribution}. 
However, this kind of integration introduces an issue that data from one dataset is not examined by the ontologies of other datasets. We will address it in steps 3 and 4.

\paragraph{Step 2: Fine-grained Definitions Generation}
\label{sec:fine-grained_definition_generation}
Fine-grained definitions of event types help models better understand the event types themselves \cite{cai-etal-2024-improving-event}. It is crucial to consider these definitions during evaluation, rather than relying solely on their names. 
Therefore, after integrating a broader ontology, we instruct GPT-4o to generate a fine-grained definition for each event type based on its event ontology and several examples.
These generated definitions average approximately 109 words, including a detailed description of the event type, an interpretation of its roles, and specific requirements derived from the examples.

\paragraph{Step 3: Latent Event Types Recognition}
To solve the problem that a data instance is not examined by the entire ontology,
we propose a cost-effective method to achieve the complete annotation in the following two steps. 
Specifically, in this step, we first instruct GPT-4o to identify possible event types and generate their corresponding fine-grained definitions. 
Then, a text similarity model \cite{bge_m3} is employed to compute pairwise similarities between possible event types and the integrated event types. 
Integrated event types that either rank in the top-$k$ for similarity with possible event types, or exceed a predefined similarity threshold, are considered as latent event types and will be used in the next step. 
In our implementation, we set $k=5$ and the threshold as 0.8.
Through this step, we filter out irrelevant event types from the text, thereby significantly reducing annotation costs. A more detailed formal discussion of the cost can be found in Appendix~\ref{sec:cost_analysis}.

\paragraph{Step 4: Supplementary Annotations}
For latent event types that may be present in the text but are not explicitly annotated, we provide their fine-grained definitions along with the incomplete data to GPT-4o for supplementary annotations. 
A full implementation example is illustrated in Appendix~\ref{sec:benchmark_construction_prompt}.
After this step, each data instance in the benchmark is completely annotated, which aligns with the characteristic of real-world open domain scenarios where each data should be examined by all possible labels.

\subsection{Benchmark Reliability Enhanced by Nucleus Sampling}
\label{sec:benchmark_reliability}
A benchmark based on LLMs generation may suffer from issues related to the reliability of annotations \cite{Chen_Qin_Jiang_Choi_2024, tan-etal-2024-large}. In the construction steps outlined above, the reliability of our benchmark is mainly influenced by the \textbf{robustness} and \textbf{quality} of the single LLM-generated annotation. Therefore, it is crucial to introduce repeated annotations and a sampling strategy to ensure the reliability of our benchmark.

Inspired by nucleus sampling in text generation \cite{holtzman2019curious}, we repeat the LLM annotation process in Figure~\ref{fig:benchmark_construction}, and calculate the frequency of each trigger across multiple annotation rounds. These triggers are then sorted by frequency in descending order, and those with a cumulative frequency reaching a threshold $p$ are selected. Next, we instruct GPT-4o to merge semantically redundant triggers. With this annotation strategy, we enable adjustable sampling by controlling the number of annotation rounds and the threshold $p$.

\begin{table}[ht]
\small
\centering
    \begin{threeparttable}
    \begin{tabular}{l|cc}
    \toprule
       Annotation Strategy  & Trg. Count & Correct Trg. (\%)\\
    \midrule
       1 round  & 310 & 80.43$_{1.59}$ \\
       3 rounds w/. $p=0.5$ & 362 & 81.12$_{1.88}$ \\
       5 rounds w/. $p=0.5$ & 389 & 83.63$_{3.65}$ \\
       10 rounds w/. $p=0.3$ & 290 & \textbf{86.90}$_{2.69}$ \\
       10 rounds w/. $p=0.5$ & 404 & 85.40$_{2.46}$ \\
       10 rounds w/. $p=0.7$ & \textbf{522} & 81.80$_{4.07}$ \\
    \bottomrule
    \end{tabular}
    \end{threeparttable}
    \caption{The results of different annotation strategies on the number of supplementary triggers (\textit{Trg. Count}) and annotation accuracy (\textit{Correct Trg.}). There is a trade-off between the diversity and accuracy of the supplementary annotations.}
    \label{tab:benchmark_reliability}
\end{table}

To validate the above strategy's effectiveness and the benchmark's reliability, we sample 200 documents (2,277 events in total) and conduct a quality check by 3 humans on the supplementary annotation data, as shown in Table~\ref{tab:benchmark_reliability}. 
Results show that (1) With a fixed $p$, increasing annotation rounds improves both the number and accuracy of supplementary triggers. This demonstrates that multiple annotations and nucleus sampling strategy are \textbf{highly effective} in enhancing the benchmark's reliability. 
(2) When the number of annotation rounds is fixed, an increase in $p$ leads to a rise in the total number of supplementary annotations, but the annotation accuracy decreases. 
This suggests that adjusting $p$ represents \textbf{a trade-off between the diversity and accuracy} of the supplementary annotations. 
A more detailed analysis can be found in Appendix~\ref{sec:reliability_analysis}. 
We will release three versions of the evaluation benchmark, based on 10 rounds of repeated annotations and using $p=0.3$, $p=0.5$, and $p=0.7$, allowing future research to choose an evaluation that favors accuracy ($p=0.3$) or diversity ($p=0.7$). In subsequent experiments, we will consistently use the balanced version ($p=0.5$) for evaluation.

\subsection{Benchmark Statistics and Scalability}
\label{sec:benchmark_analysis}
After multiple rounds of supplementary annotation and nucleus sampling, the number of events in the benchmark increases by a factor of 2.29 compared to the original data. More detailed statistics can be found in the Appendix~\ref{sec:statistics}.

Since ODED aims to detect all possible events from any potential domain, it is essential to ensure the scalability of the ODED evaluation benchmark.  
In SEOE, both data (i.e., event-containing documents) and event types are scalable. Specifically, when new data is added to the benchmark, step 3 and 4 are repeated to achieve new annotations. When new event types are added, step 2 generates their fine-grained definitions. Then, step 3 and 4 perform new supplementary annotations for previous data.
Since step 3 employs a text similarity model for filtering, the additional cost incurred from adding new data and event types is approximately linear with the number of data instances and event types. This ensures the benchmark's cost-effective scalability and enhances the model's representativeness of real-world scenarios. To support this conclusion, a formulaic analysis can be found in Appendix~\ref{sec:scalability_analysis}.



\section{Semantic Evaluation by LLMs}
With a clear task definition and a representative evaluation benchmark, we will introduce how to achieve semantic evaluation for ODED. Section~\ref{sec:groups} includes a proposed module to assist in evaluation. Section~\ref{sec:evaluation_process} includes how to apply LLMs as automatic evaluation agents to compute a semantic F1-score. Finally, we will show that GPT-4o's evaluation has a strong correlation with human judgments in Section~\ref{sec:correlation}.

\subsection{Semantically Similar Fine-grained Definitions as Groups}
\label{sec:groups}
\begin{figure}[h]
    \centering
    \includegraphics[width=0.9\linewidth]{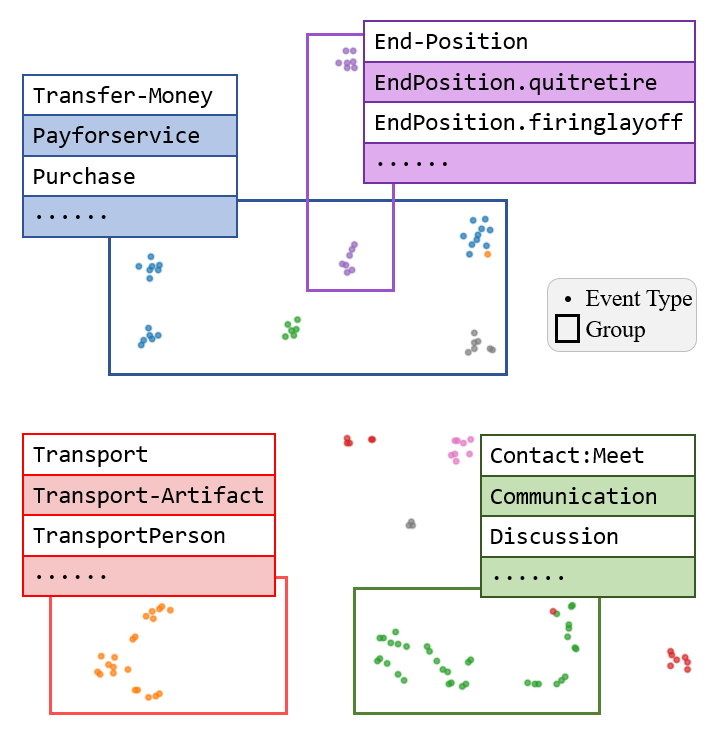}
    \caption{T-SNE \cite{JMLR:v9:vandermaaten08a} visualization results of our constructed groups. To maintain conciseness, we only present a subset of groups.}
    \label{fig:group_illustration}
\end{figure}

Previous work validates that ontology information, such as relationships among event types, helps models with the understanding of the event, and generalize better to unseen event types \cite{cai-etal-2024-improving-event}. 
Since our integrated ontology lacks such information, we propose a definition-aware method to construct it. 
Specifically, we leverage fine-grained definitions generated from Section~\ref{sec:fine-grained_definition_generation} to compute the similarity between each pair of event types by applying an off-the-shelf text similarity model \cite{bge_m3}. 
Event type pairs with similarity exceeding a predefined threshold are considered highly related, and all such related event types are regarded as a \textbf{group}, as illustrated in Figure~\ref{fig:group_illustration}. 
In the following evaluation process, we will provide all event types within the group of the evaluated event type as the ontology information. The effectiveness of such group module will be validated in Section~\ref{sec:correlation}.

\subsection{Semantic Evaluation Process}
\label{sec:evaluation_process}
We first formalize the evaluation task for ODED. Suggest there is a tuple $\{T, {\bf P}, {\bf D}_p, {\bf G}, {\bf D}_g\}$, where $T$ is an extracted text, $\bf P$ is a predicted trigger set from an open ED model, ${\bf D}_p$ is the predicted event definition set, $\bf G$ is a golden trigger set from a benchmark, and ${\bf D}_g$ is the golden event definition set.
An ODED evaluation framework is required to identify the correctness of each prediction in $\bf P$ and the recall of each annotation in $\bf G$.

In SEOE, similar to \cite{lu2024exactmatchsemanticallyreassessing}, we utilize LLMs as the automatic semantic evaluation agent. The above tuple along with corresponding group information as mentioned in Section~\ref{sec:groups} will be prompted to the agent. 
The agent is required to identify a semantic correspondence set $\bf C$.
The elements in $\bf C$, denoted as $(p,g)$, where $p\in {\bf P},g \in {\bf G}$, indicate that prediction $p$ and golden annotation $g$ are semantically matched.
Then, semantic-level F1 score can be computed as follows:
\begin{align*}
{\bf C_p} &= \{p \mid  (p,g) \in {\bf C}\}, \\
\quad {\bf C_g} &= \{g \mid (p,g) \in {\bf C}\}, \\
\bf p = \frac{|C_p|}{|P|},\quad r &= \bf \frac{|C_g|}{|G|},\quad F1 = \frac{2*p*r}{p+r}
\end{align*}



\subsection{LLMs' Evaluation Correlation with Human Judgments}
\label{sec:correlation}
To prove the effectiveness of our proposed SEOE framework and several modules, we conduct a meta-evaluation to analyze the correlation between SEOE and human judgments. Specifically, we invite 3 professional event annotators to evaluate 200 event-containing documents (791 prediction-gold event pairs in total). These annotators are given the same prompts as those given to LLMs. 
As for the evaluation agents, we select the top-3 LLMs that demonstrate superior performance in text evaluation\footnote{https://openai.com/index/hello-gpt-4o/}.
As for the evaluation metrics, we compute three widely-used inter-annotator agreement (IAA) metrics among humans and LLMs: percent agreement \cite{zheng2023judging, liu2021towards}, Spearman correlation \cite{lan2024criticbench}, and Cohen's Kappa \cite{thakur2024judging}. To minimize the impact of randomness from LLMs, 
we repeat the experiment three times and report the average and standard deviation of the results, as shown in Table~\ref{tab:correlation_with_human}.

\begin{table}[h]
\small
\centering
\begin{threeparttable}
    \begin{tabular}{l|ccc}
        \toprule
          Comparison  & Percent & Spearman  & Cohen's \\
        \midrule
          3 Humans  & 95.32 & 79.92 & 79.54 \\
        \midrule
           GPT-4o & \textbf{94.41}$_{0.14}$ & \textbf{77.50}$_{0.39}$ & \textbf{77.03}$_{0.44}$ \\ 
           ~~ - w/o. Groups & 93.37$_{0.20}$ & 74.63$_{0.52}$ & 73.73$_{0.61}$ \\
           ~~ - w/o. \#Defs & 92.97$_{0.28}$ & 73.85$_{0.94}$ & 72.67$_{1.00}$ \\
           Claude3-Opus & 93.95$_{0.03}$ & 76.03$_{0.35}$ & 75.43$_{0.27}$ \\
           GPT-4-turbo & 93.78$_{0.09}$ & 74.99$_{0.54}$ & 74.50$_{0.49}$ \\
        \bottomrule
        \end{tabular}
    \caption{Three inter-annotator agreement (IAA) metrics (\%) among humans and LLMs. \textit{\#Defs}: Fine-grained Definitions. \textit{3 Humans} reports the IAA among humans, regarded as the golden standard. IAA between a LLM and each human evaluator is calculated first, and then the average value is taken.  All Spearman correlation scores are associated with the $p$-value $<$ 0.005.}
    \label{tab:correlation_with_human}
\end{threeparttable}
\end{table}

According to Table~\ref{tab:correlation_with_human}, it is observed that GPT-4o outperforms other LLMs in terms of IAA with human evaluation. Its evaluation results are highly aligned with pure human assessments, with a percent agreement difference of 0.91\%, a Spearman correlation difference of 2.42\%, and a Cohen's Kappa difference of 2.51\%. 
Furthermore, the introduction of group information and fine-grained definitions significantly improves IAA among GPT-4o and humans. 
We will use GPT-4o as the automatic evaluation agent in the subsequent evaluation experiments.

\section{Evaluation Experiments}
With a complete and widely domain-covered benchmark along with a reliable evaluation process, we will evaluate existing ODED models in this section. Apart from the main evaluation results in Section~\ref{sec:main_results}, we will analyze patterns of incorrect predictions of evaluated ODED models in Section~\ref{sec:incorrect_analysis}.

\begin{table}[ht]
\small
\centering
\begin{threeparttable}
    \begin{tabular}{l|ccc}
    \toprule
       \multirow{2}{*}{\bf Models}  & \multicolumn{3}{c}{\textbf{Trigger Classification (\%)}} \\ \cline{2-4}
       ~ & \textbf{precision} & \textbf{recall} & \textbf{F1-score} \\
    \midrule
    \multicolumn{4}{l}{\textbf{\textit{Closed-source GPT Series LLMs}}} \\ 
    \midrule
        \textbf{o1-preview} & 62.67 & \textbf{56.59} & \textbf{59.47} \\
        \textbf{GPT-4} & 67.46 & 45.12 & 54.08 \\
        \textbf{GPT-4-turbo} & 66.53 & 45.23 & 53.85 \\
        \textbf{GPT-4o} & 73.58 & 45.61 & 56.31 \\
        \textbf{GPT-3.5-turbo} & 69.43 & 42.89 & 53.02 \\
    \midrule
    \multicolumn{4}{l}{\textbf{\textit{Closed-source Claude Series LLMs}}} \\ 
    \midrule
        \textbf{Claude-3.5-haiku} & \textbf{75.35} & 44.26 & 55.76 \\
        \textbf{Claude-3.5-sonnet} & 68.47 & 48.20 & 56.57 \\
    \midrule
    \multicolumn{4}{l}{\textbf{\textit{Closed-source Gemini Series LLMs}}} \\ 
    \midrule
        \textbf{Gemini-1.5-flash} & 64.68 & 52.41 & 57.90 \\
        \textbf{Gemini-1.5-pro} & 63.45 & 49.19 & 55.42 \\
    \midrule
    \multicolumn{4}{l}{\textbf{\textit{Open-source Qwen Series LLMs}} \cite{bai2024benchmarking}} \\ 
    \midrule
        \textbf{Qwen-2.5-72B-Chat} & 68.83 & \textbf{49.20} & \textbf{57.38} \\
        \textbf{Qwen-2.5-7B-Chat} & 71.16 & 43.43 & 53.94 \\
    \midrule
    \multicolumn{4}{l}{\textbf{\textit{Open-source InternLM Series LLMs}} \cite{team2023internlm}} \\ 
    \midrule
        \textbf{InternLM-2.5-20B} & 72.08 & 43.84 & 54.52 \\  
        \textbf{InternLM-2.5-7B} & 71.64 & 42.88 & 53.65 \\
    \midrule
    \multicolumn{4}{l}{\textbf{\textit{Open-source Llama-3 Series LLMs}} \cite{dubey2024llama}} \\ 
    \midrule
        \textbf{Llama-3.2-70B-Chat} & 72.28 & 45.77 & 56.05 \\
        \textbf{Llama-3.1-8B-Chat} & \textbf{74.77} & 42.73 & 54.38 \\
    \bottomrule
    \end{tabular}
    \caption{Closed-source and open-source LLMs' open domain event detection results evaluated by SEOE.}
    \label{tab:main_results}
\end{threeparttable}
\end{table}

\subsection{Experimental Settings}
Following the ODED task formulation in Section~\ref{sec:task_formulation}, it is challenging for small language models to identify cases of unseen event types \cite{veyseh2021augmenting} along with fine-grained definitions. Therefore, we only conduct evaluation experiments on existing closed-source and open-source LLMs.

To mitigate the impact of randomness caused by the stochastic process in LLMs \cite{bouras2024integratingrandomnesslargelanguage}, we will use deterministic decoding by setting the temperature parameter of all evaluated LLMs to 0.



\subsection{Overall Analysis of Evaluation Results}
\label{sec:main_results}
Table~\ref{tab:main_results} shows LLMs' ODED results evaluated by SEOE, indicating several findings: (1) Closed-source LLMs slightly outperform open-source LLMs on the ODED task. Among the closed-source models, o1-preview achieves the best performance with an F1-score of 59.47. The best-performing open-source model is Qwen-2.5-72B, with an F1-score of 57.38. 
(2) The results from open-source models reveal a clear trend: as the number of model parameters increases, the model's F1-score improves, with this improvement primarily driven by gains in recall. 
(3) Across all LLMs, the F1-scores are mainly affected by relatively low recall rather than precision, indicating that while LLMs ensure the accuracy of extracted event triggers, they may lack the capability to comprehensively identify a broader range of triggers. 

To further validate statement (3), we select the top-3 LLMs with the highest recall scores and the bottom-3 LLMs with the lowest recall scores from Table~\ref{tab:main_results}, and calculate their average number of predicted triggers, as shown in Table~\ref{tab:recall_analysis}. The results indicate a positive correlation between recall and the average number of predicted triggers, and a negative correlation with precision. It further reveals a trade-off between \textbf{accuracy} and \textbf{diversity} in model predictions for the ODED task. Besides, this phenomenon is highly similar to the findings discussed in Section~\ref{sec:benchmark_reliability}, suggesting that ODED models may improve the quality of open domain extraction results through multiple extractions with high-quality sampling \cite{holtzman2019curious}.

\begin{table}[ht]
\small
\centering
\begin{threeparttable}
    \begin{tabular}{l|ccc}
    \toprule
       \textbf{Models}  & \textbf{precision} & \textbf{recall} & \textbf{\#TrgNum} \\
    \midrule
       \textbf{o1-preview}  & 62.67 & \textbf{56.58} & \textbf{4.59} \\
       \textbf{Gemini-1.5-flash} & 64.68 & 52.41 & 4.17 \\
       \textbf{Qwen-2.5-72B-Chat} & 68.83 & 49.20 & 3.59 \\
    \midrule
       15 LLMs in Average  & -- & -- & 3.39 \\
    \midrule
       \textbf{GPT-3.5-turbo}  & 69.43 & 42.89 & 3.02 \\
       \textbf{InternLM-2.5-7B} & 71.64 & 42.88 & 2.81 \\
       \textbf{Llama-3.1-8B-Chat} & \textbf{74.77} & 42.73 & 2.74 \\
    \bottomrule
    \end{tabular}
    \caption{Top-3 and bottom-3 recall score LLMs' average number of predicted triggers (\textit{\#TrgNum}).}
    \label{tab:recall_analysis}
\end{threeparttable}
\end{table}

\subsection{Incorrect Predictions Analysis}
\label{sec:incorrect_analysis}
The main characteristic of ODED is that models are not restricted to predefined event types during extraction, and are required to generate event types with definitions. Therefore, the error classification in closed domain \cite{Chen_Qin_Jiang_Choi_2024, lu2024exactmatchsemanticallyreassessing} is not applicable to analyze incorrect predictions in ODED. Based on the observation of human quality check in Section~\ref{sec:benchmark_reliability}, we ascribe incorrect predictions to several patterns, including:


\begin{itemize}
    \item \textbf{Ambiguous Mention}: The mention of prediction is too brief, leading to unclear semantics, making it difficult to determine the event it refers to.
    \item \textbf{Lengthy Mention}: The mention of prediction is too lengthy, complicating the accurate identification of a certain event.
    \item \textbf{Inconsistent Type Definition}: With a semantic golden match to the mention, the definition of predicted type is overly broad or too narrow compared to that of the golden match.
    \item \textbf{Conflicting Type Definition}: With a semantic golden match to the mention, the definition of predicted type has conflicts with the definition of the golden match.
    \item \textbf{Reasonable Prediction with No Match}: The prediction can be considered reasonable but lacks a semantic golden match.
\end{itemize}

\begin{figure}[htbp]
    \centering
    \begin{subfigure}[t]{0.45\textwidth}
        \centering
        \includegraphics[width=\textwidth]{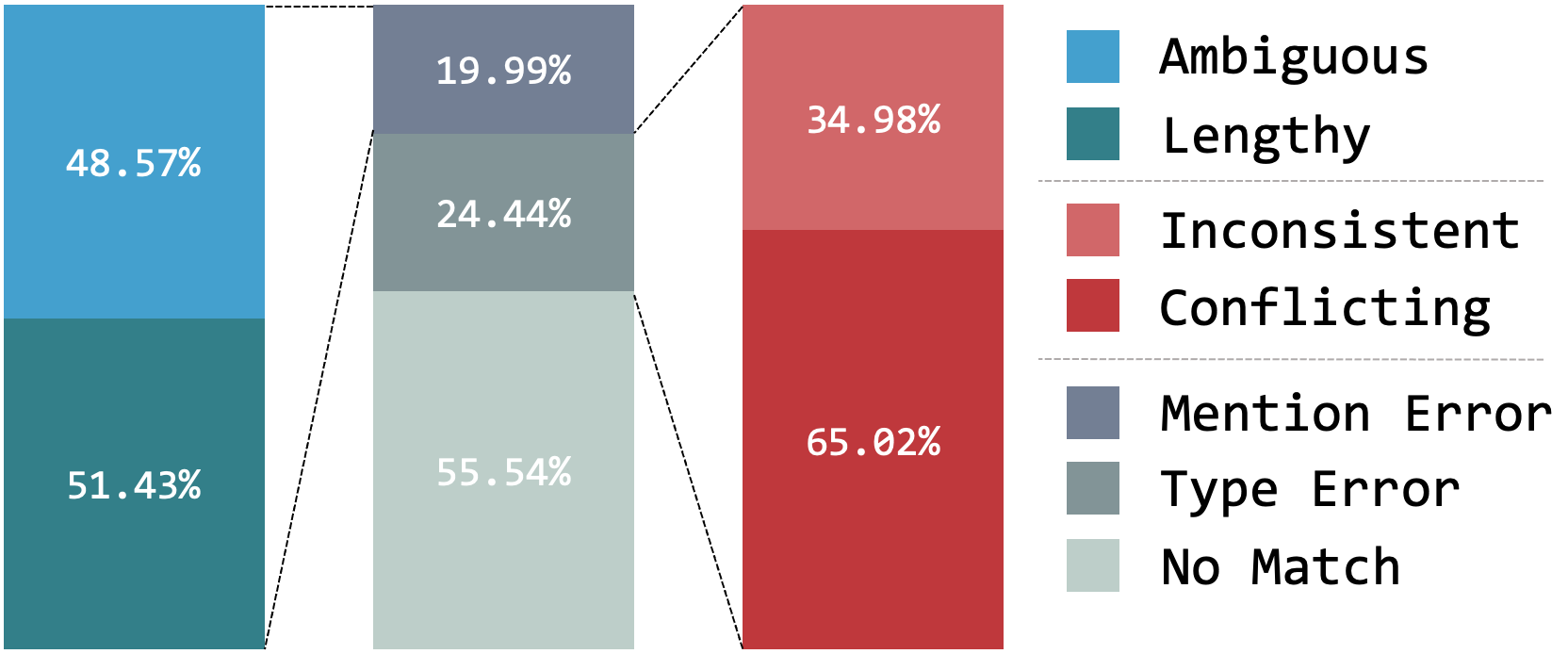}
        \caption{Overall error pattern distribution.}
        \label{fig:subfig_a}
    \end{subfigure}
    \\
    \begin{subfigure}[t]{0.45\textwidth}
        \centering
        \includegraphics[width=\textwidth]{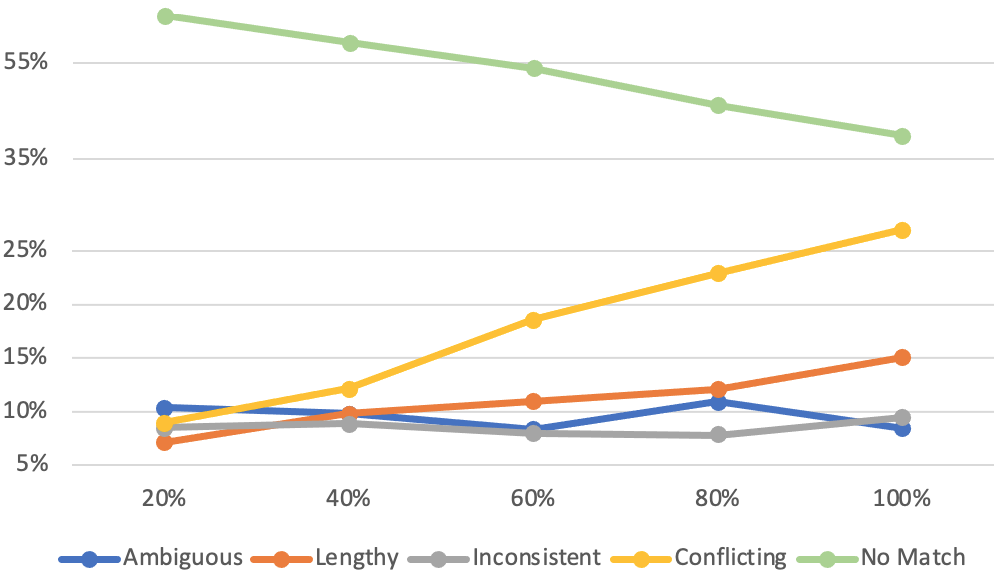}
        \caption{The trend of error distribution with the variation in event density across texts. The event density is defined as the number of events in a text divided by the text length. We sort the data based on event density and then present error distribution at different cumulative rankings.}
        \label{fig:subfig_b}
    \end{subfigure}
    \caption{Error Pattern Distribution. All data are randomly sampled from the main experiment.}
    \label{fig:incorrect_distribution}
\end{figure}

We manually annotate the distribution of the above error patterns of the incorrect predictions across the entire benchmark, as shown in Figure~\ref{fig:incorrect_distribution}.  
Overall, the most frequent error type is \textit{No Match} (55.54\%), and its proportion decreases as the event density of the text increases. This phenomenon can be interpreted in two aspects: (1) it suggests that the ODED benchmark may be lacking certain event types, highlighting the need for the benchmark's scalability; (2) since the benchmark's ontology is derived from mainstream event extraction research, it indicates that a significant proportion of events extracted by the evaluated ODED models are not covered by existing event extraction studies, further suggesting that the extraction strategy of ODED models needs targeted adjustments.


Additionally, overall, the proportion of errors in the \textit{type} category (24.44\%) exceeds that in the \textit{mention} category (19.99\%), and this gap is even more pronounced in documents with higher event density. It indicates that compared to extracting mentions, determining the type of a mention and defining the type is a more challenging task in ODED. 



\section{Conclusion and Future Works}
In this paper, we propose a scalable and reliable ODED evaluation framework, named SEOE. It consists of a more representative benchmark and a well-validated semantic evaluation process. Experiments reveal that ODED is a challenging task, especially in predicting event types and corresponding definitions. In the future, we will explore the evaluation of open domain event argument extraction and investigate methods to enhance open domain event extraction.

\section*{Limitations}
\paragraph{LLMs are not Perfect Human Substitute} 
In this paper, advanced LLMs are used as evaluation agents, showing very strong correlation with human judgments. However, it is important to note that LLMs cannot perfectly align with human evaluations. In the future, we would like discover other strategies to further improve the reliability of LLM-as-a-Judge, such as multi-agent debate \cite{chan2023chatevalbetterllmbasedevaluators}.

\paragraph{Correctness of Generated Definitions}
In our annotation and evaluation experiments, GPT-4o is used to generate fine-grained definitions for all event types to assist with supplementary annotation and semantic evaluation. While experiments have demonstrated the effectiveness of this module, it is important to note that these fine-grained definitions might have errors.

\paragraph{Cost of LLM-based Evaluation}
Unlike evaluations based on token-level matching rules, our evaluation framework utilizes advanced LLMs to assess models' ODED predictions, which introduces higher costs. In the case of the full benchmark, which contains 4,423 documents, the cost of evaluating a model is approximately \$59.8.
However, it is possible to accurately evaluate models using only a subset of the benchmark, as the average frequency of each event type in our benchmark is approximately 50.8, which may exceed the requirements for model evaluation.
To validate this hypothesis, we identify subsets of the benchmark with sizes of 1/10, 1/4, and 1/2 through random sampling. The average differences in evaluation metrics between these subsets and the full benchmark used in the main experiments are 1.06\%, 0.57\%, and 0.38\%, respectively. This indicates that using an appropriately chosen subset can significantly reduce evaluation costs while maintaining statistical representativeness.
These subsets of the benchmark will also be publicly available.

\paragraph{Evaluation without Offsets}
In our proposed evaluation framework, we do not apply offsets to the golden annotations in the benchmark, nor do we require the evaluated models to output the corresponding offsets for their predictions. This setting might introduce ambiguity when events with the same mention and event type appear at different positions in the text. 
Our evaluation decision is based on three main considerations: (1) Adding offsets to the generated results is a common challenge in generative extraction tasks. Some studies \cite{lu-etal-2022-unified} have shown that this issue occurs in a small proportion of cases and can be addressed using simple heuristics, such as mapping spans to offsets by finding the first matched offsets, which results in minimal errors (<0.5\%). (2) The ability of ODED models to output offsets is not sufficiently reliable, and requiring them to do so could lead to less accurate evaluation results. (3) The proposed semantic-level evaluation does not depend on tokens, so the need for offsets is minimal. We plan to explore the specific impact of offsets on ODED evaluation in future work.

\bibliography{main}

\appendix

\section{Benchmark Integration}
\label{sec:benchmark_integration}
We utilize a unified event extraction framework \cite{huang-etal-2024-textee} to perform the integration. Basic statistics about sampled datasets are presented in Table~\ref{tab:integration_statistics}.

During the integration, we uniformly sample cases across all types. Specifically, during the sampling process, we try to ensure that each type appears in at least 10 documents. Once a type appears in 10 documents, we stop sampling it further.

We do not merge event types with the same name from different datasets originating from the same ontology. This is because different datasets may have their own customized annotation or interpretation rules for the same ontology, which are inflected in their distinctive annotation guidelines. 
We add serial numbers to distinguish event types with the same name.
This treatment also enhances the feasibility of scaling our benchmark.

\begin{table*}[h]
\centering
\small
\begin{threeparttable}
    \begin{tabular}{l|crrrr}
    \toprule
       Test Dataset / Benchmark & Domain & Documents & Tokens & Event Types & Events \\
    \midrule
       ACE05 \cite{doddington-etal-2004-automatic} & News & 2519 & 33824 & 30 & 563 \\
       MUC-4 \cite{muc1992muc4}  & News & 464 & 120751 & 1 & 464 \\
       RAMS  \cite{ebner-etal-2020-multi} & News & 910 & 120162 & 135 & 910 \\
       PHEE  \cite{sun-etal-2022-phee} & Pharmacovigilance & 965 & 21377 & 2 & 1005 \\
       Genia2013 \cite{kim-etal-2013-genia} & Biomedical & 139 & 31341 & 11 & 974 \\
       MLEE  \cite{Pyysalo12mlee} & Biomedical & 42 & 10440 & 21 & 867\\
       CASIE \cite{Satyapanich20casie} & Cybersecurity & 218 & 63103 & 5 & 1268 \\
       FewEvent \cite{Deng20fewevent} & General & 2541 & 80698 & 99 & 2541 \\
       MAVEN \cite{wang-etal-2020-maven} & General & \textbf{5925} & 150744 & 168 & 14190 \\
       M2E2 \cite{li-etal-2020-cross} & Multimedia & 901 & 25467 & 8 & 174 \\
       MEE-en \cite{pouran-ben-veyseh-etal-2022-mee} & Wikipedia & 1300 & 106320 & 16 & 1745 \\
       WikiEvents \cite{li-etal-2021-document} & Wikipedia & 62 & 21010 & 38 & 379 \\
    \midrule
       SEOE (Ours, with $p = 0.5$) & Open Domain & 4423 & \textbf{382667} & \textbf{564} & \textbf{28653} \\
    \bottomrule
    \end{tabular}
    \caption{Basic statistics about sampled datasets used in SEOE. Since our proposed SEOE is an evaluation dataset, we only select the test sets of other datasets for comparison.}
    \label{tab:integration_statistics}
\end{threeparttable}
\end{table*}

\section{Benchmark Statistics}
\label{sec:statistics}
Table~\ref{tab:statistics} presents basic statistics of 3 different versions of our proposed SEOE's benchmark. The benchmark with $p=0.3$ focuses more on evaluating the accuracy of the model's extraction results, while the benchmark with $p=0.7$ emphasizes the evaluation of the model's extraction diversity. The benchmark with $p=0.5$ is a balanced version and is used in the evaluation experiments presented in the main text.

We also provide a comparison of event type distribution in SEOE and other datasets, shown in  Table~\ref{tab:type_distribution}.

\begin{table}[ht]
\small
\centering
\begin{threeparttable}
    \begin{tabular}{l|rrr}
    \toprule
       Metric & $p=0.3$ & $p=0.5$ & $p=0.7$ \\
    \midrule
       Documents  & 4423 & -- & -- \\
       Sentences & 15095 & -- & -- \\
       Tokens & 382667 & -- & -- \\
       Event Types & 564 & -- & -- \\
       Avg. Def. Length & 109.03 & -- & -- \\
       Events & 22333 & 28653 & 35015 \\
       Avg. Trigger Length & 1.27 & 1.42 & 1.53 \\
       Avg. Events per Doc. & 5.04 & 6.47 & 7.91 \\  
       Type's Avg. Frequency & 39.60 & 50.80 & 62.08 \\ 
    \bottomrule
    \end{tabular}
    \caption{Statistics of SEOE's benchmark. The benchmark's supplemental annotations are derived from 10 independent repetitions of annotations along with a nucleus sampling with different threshold $p$.}
    \label{tab:statistics}
\end{threeparttable}
\end{table}

\begin{table*}[ht]
\small
\centering
\begin{threeparttable}
    
    \begin{tabular}{l|rrrrrr}
    \toprule
       Test Dataset / Benchmark  & Event Type & Top-10\% & Top-25\% & Bottom-50\% & Bottom-25\% & Bottom-10\%  \\
    \midrule
       SEOE (Ours, with $p = 0.3$) & 564 & 37.39 & 59.54 & 19.65 & \textbf{7.26} & \textbf{2.22} \\
       SEOE (Ours, with $p = 0.5$) & 564 & 36.10 & 59.36 & 19.08 & 6.63 & 1.89 \\
       SEOE (Ours, with $p = 0.7$) & 564 & 35.94 & 59.89 & 18.65 & 6.30 & 1.81 \\
    \midrule
       MAVEN  & 168 & 40.09 & 67.13 &  11.21 & 2.67 & 0.40 \\
       RAMS  & 135 & 34.28 & \textbf{55.49} & \textbf{20.21} & 5.71 & 1.42 \\
       FewEvent  & 99 & \textbf{21.25} & 56.67 & 10.74 & 4.64 & 1.18 \\
       WikiEvents  & 38 & 37.46 & 69.39 & 10.29 & 2.37 & 0.79 \\
       ACE05  & 30 & 58.43 & 79.21 & 6.39 & 1.59 & 0.53 \\
       MLEE  & 21 & 32.75 & 68.05 & 7.61 & 1.96 & 0.46 \\
    \bottomrule         
    \end{tabular}
    \caption{Cumulative distribution (\%) of event types at different frequency levels. Top-$x$\% and Bottom-$x$\% refer to the total proportion of event types ranked in the top x\% and bottom x\% based on frequency, respectively. Datasets only with the number of event types over 20 are selected for comparison. Since our proposed SEOE is an evaluation dataset, we only select the test sets of other datasets for comparison.}
    \label{tab:type_distribution}
\end{threeparttable}
\end{table*}

\section{Benchmark Reliability Analysis}
\label{sec:reliability_analysis}
In the main text, we report the annotation accuracy under different annotation strategies, using only the classifications of\textit{ correct} or \textit{incorrect}. In fact, during the annotation process, we further subdivide these two types into the following seven subtypes:
\begin{itemize}
    \item C1: Nearly perfect annotations.
    \item C2: Semantically correct, but the mention is slightly longer than necessary.
    \item C3: Semantically correct, but the event type is slightly too broad.
    \item C4: Semantically correct, but semantically redundant with one of the golden annotations.
    \item IC1: Incorrect annotation due to an incorrect mention.
    \item IC2: Incorrect annotation due to an incorrect event type.
    \item IC3: Completely incorrect annotation at both the mention and event type levels, or the selection of words not present in the original text.
\end{itemize}

The proportion of the above 7 subtypes under different annotation strategies is illustrated in Table~\ref{tab:reliability_correct} and \ref{tab:reliability_incorrect}.
It is worth noting that, although we use GPT-4o to merge semantically redundant annotations after the nucleus sampling, as mentioned in the main text, a substantial proportion of semantically redundant annotations still remain (i.e., \textit{C4} in Table~\ref{tab:reliability_correct}). Further observation reveals that this redundancy is primarily caused by two main factors: (1) similar events occurring at different positions in the text, and (2) different perspectives on the same event, such as the text \textit{A mother gave birth to a baby}, which might involve both a \textit{Life:Pregnancy} event and a \textit{Life:Born} event, leading to semantic overlap. Nevertheless, these two factors do not affect the accuracy of the subsequent evaluation, and therefore we still consider these annotations as correct.

\begin{table}[ht]
\small
\centering
    \begin{threeparttable}
    \begin{tabular}{l|cccc}
    \toprule
       Annotation Strategy  & C1 & C2 & C3 & C4 \\
    \midrule
       1 round  & 52.37 & 2.26 & 1.40 & 24.41 \\
       3 rounds w/. $p=0.5$ &  50.37 & 4.60 & 1.38 & 24.77 \\
       5 rounds w/. $p=0.5$ & 49.70  & 4.37 & 1.71 & 27.85 \\
       10 rounds w/. $p=0.3$ & \textbf{54.60}  & 3.56 & \textbf{1.84} & 26.90  \\
       10 rounds w/. $p=0.5$ & 49.75  & 5.03  & 1.24 & 29.37  \\
       10 rounds w/. $p=0.7$ & 44.13  & \textbf{5.62} & 1.15 & \textbf{30.91} \\
    \bottomrule
    \end{tabular}
    \end{threeparttable}
    \caption{The proportion (\%) of annotations classified into each correct subtype under different annotation strategies, relative to the total number of annotations.}
    \label{tab:reliability_correct}
\end{table}

\begin{table}[ht]
\small
\centering
    \begin{threeparttable}
    \begin{tabular}{l|ccc}
    \toprule
       Annotation Strategy  & IC1 & IC2 & IC3 \\
    \midrule
       1 round   & 2.80 & \textbf{14.95} & 1.83  \\
       3 rounds w/. $p=0.5$   & 2.85 & 14.27 & 1.75 \\
       5 rounds w/. $p=0.5$   & 3.17 & 11.74 & 1.46 \\
       10 rounds w/. $p=0.3$  & 1.61 & 9.43 & 2.07  \\
       10 rounds w/. $p=0.5$  & 2.39  & 10.56 & 1.65  \\
       10 rounds w/. $p=0.7$   & \textbf{3.70} & 11.43 & \textbf{3.07}  \\
    \bottomrule
    \end{tabular}
    \end{threeparttable}
    \caption{The proportion (\%) of annotations classified into each incorrect subtype under different annotation strategies, relative to the total number of annotations.}
    \label{tab:reliability_incorrect}
\end{table}


\section{Benchmark Cost and Scalability Analysis}
Our proposed cost-effective method achieves complete annotations for constructing an open domain event detection evaluation benchmark. We will analyze how this method reduces the cost of benchmark construction and facilitates the scalability of the benchmark. 

\subsection{Cost Analysis}
\label{sec:cost_analysis}
Assuming that there are $N$ texts and $T$ event types in the benchmark, then the naive annotation cost is 
$$
C_{naive}(N,T) = N \times T \times \Bar{c}
$$
, where $\Bar{c}$ represents the average cost of a single annotation of humans or agents such as LLMs. 

In our construction pipelines, LLMs are required to identify possible event types in the texts, with an average of $\Bar{p}$ types per text. Then, a single-tower model for text similarity is employed to filter out latent event types, with an average of $\Bar{t}$ types, from the benchmark that exhibit high similarity to possible event types. Therefore, the cost-effective annotation cost is 
$$
C(N,T) = N \times \Bar{t} \times \Bar{c} + N \times \Bar{c} + (N \times \Bar{p} + T) \times \Bar{c'}
$$
, where $\Bar{c'}$ represents the average cost of computing a text embedding with the single-tower model, and $\Bar{t} \approx 10 \ll T $. The overall cost arises from three parts: (1) performing supplementary annotation with $\Bar{t}$ latent event types, (2) detecting possible event types for each text, and (3) computing embeddings for $T$ integrated event types and $\Bar{p}$ possible of each text using the single-tower model.
As the single-tower model is typically derived from an off-the-shelf small language model, the third part of the cost is negligible. Therefore, the overall cost can be estimated as follows:
$$
C(N,T) \approx N \times \Bar{t} \times \Bar{c} \ll C_{naive}(N,T)
$$

The above analysis demonstrates that our proposed cost-effective method can reduce annotation costs to nearly linear scaling with the number of texts when applied to benchmarks with extensive text data and numerous types. 

\subsection{Scalability Analysis}
\label{sec:scalability_analysis}
Assuming that there are $N'$ new texts and $T'$ new types as the supplement of the benchmark, there are three components of additional cost. The first cost comes from generating fine-grained definitions for new event types, which is linearly proportional to the number of new types:
$$
\Delta C_1 = T' \times \Bar{c_t}
$$
, where $\Bar{c_t}$ represents the average cost of generate the fine-grained definition for a new event type.

The second cost involves annotating new texts using the newly integrated types, which is similar to Section~\ref{sec:cost_analysis}:
\begin{align*}
\Delta C_2 &= N' \times (\Bar{t} + \Bar{t'} + 1) \times \Bar{c} \\
& \quad + (N' \times \Bar{p} + T') \times \Bar{c'} \\
& \approx N' \times (\Bar{t} + \Bar{t'}) \times \Bar{c}
\end{align*}
, where $\Bar{t'} \ll T'$ denotes new latent type from newly added types. This part of the cost is approximately linearly proportional to the number of new texts.

The third cost arises from supplementing the annotated $T$ texts with new types:
$$
\Delta C_3 = N \times \Bar{t'} \times \Bar{c}
$$ 

Let $N_{all} = N + N' $. Then, the total extra cost is:
\begin{align*}
\Delta C(N',T') &= \Delta C_1 + \Delta C_2 + \Delta C_3 \\
&\approx T' \times \Bar{c_t} + N_{all} \times \Bar{t'} \times \Bar{c} \\
&+ N' \times \Bar{t} \times \Bar{c}
\end{align*}

The above analysis demonstrates that when new texts and types are added to the benchmark, the additional cost is primarily linearly proportional to the number of new types, new texts and annotated texts that need additional annotation. In conclusion, our proposed cost-effective method supports the open benchmark easily scalable.




\section{Prompts}
\subsection{Benchmark Construction Prompts}
\label{sec:benchmark_construction_prompt}
In Section~\ref{sec:benchmark}, we leverage GPT-4o to construct the evaluation benchmark. The prompt used for generating fine-grained definitions for each event type is illustrated with an example in Table~\ref{tab:generate_definitions_prompt}. The prompt used for identifying possible event types for each data instance is illustrated with an example in Table~\ref{tab:identify_possible_event_types_prompt}. After recognizing latent event types, the prompt used for generating supplementary annotations is illustrated with an example in Table~\ref{tab:generate_supplementary_annotations_prompt}. After nucleus sampling from multiple repeated annotations, the prompt used for merging semantically redundant triggers is illustrated with an example in Table~\ref{tab:merge_redundant_triggers_prompt}.

\begin{table*}[h]
    \centering
    \small
    \begin{threeparttable}
    \begin{tabularx}{\textwidth}{cX}
     \toprule
        \multicolumn{2}{l}{\textbf{Prompt Used for Generating Fine-grained Definitions for Each Event Type}} \\
     \midrule
        \multirow{3}{*}{Background} & In the context of open domain event extraction, we aim to enhance our understanding of various event types by generating more detailed and fine-grained definitions. This will aid in better classification and recognition of events in text. \\
        ~ \\
        \multirow{5}{*}{Task Description} & You are a knowledgeable event extraction specialist tasked with generating a more detailed definition for a specified event type based on its name, ontology, and provided examples of this event type. Your definition should capture the nuances and variations of the event type to facilitate its application in natural language processing tasks. Input: You will receive an event type along with its ontology. Some examples of this event type are provided.\\
        ~ \\
        \multirow{2}{*}{Output Format} & Please provide the output in the following Python dictionary format: \{"event\_type": <str: given\_event\_type\_name>, "detailed\_definition": <str: your\_generated\_definition>\}  \\
        ~ \\
        \multirow{19}{*}{Data} & Event Type Name: Conflict:Attack \\
        ~ & Ontology: The argument roles of this event type are ['Attacker', 'Target', 'Instrument', 'Place', 'Agent']. \\ 
        ~ & Example 0: \\
        ~ & \quad Text: This isn't a football game, and it's not two drunken cousins [t] fighting [/t] at a family reunion. \\
        ~ & \quad Trigger: fighting \# event type: Conflict:Attack \\
        ~ & \quad Argument 0: cousins \# role: Attacker \\
        ~ & \quad Argument 1: cousins \# role: Target \\
        ~ & Example 1: \\
        ~ & \quad Text: When we came across the bridge, the enemy [t] launched [/t] some 155 artillery rounds at us. \\
        ~ & \quad Trigger: launched \# event type: Conflict:Attack \\
        ~ & \quad Argument 0: enemy \# role: Attacker \\
        ~ & \quad Argument 1: bridge \# role: Place \\
        ~ & \quad Argument 2: us \# role: Target \\
        ~ & \quad Argument 3: rounds \# role: Instrument \\
        ~ & Example 2: \\
        ~ & \quad Text: luster skipped bail during his trial in January for [t] raping and drugging [/t] three women. \\
        ~ & \quad Trigger: raping and drugging \# event type: Conflict:Attack \\
        ~ & \quad Argument 0: women \# role: Victim \\
        ~ & \quad Argument 1: luster \# role: Agent \\
     \bottomrule
    \end{tabularx}
    \end{threeparttable}
    \caption{Prompt used for generating fine-grained definitions for each event type in the integrated ontology. An example of data is illustrated.}
    \label{tab:generate_definitions_prompt}
\end{table*}

\begin{table*}[h]
    \centering
    \small
    \begin{threeparttable}
    \begin{tabularx}{\textwidth}{cX}
     \toprule
        \multicolumn{2}{l}{\textbf{Prompt Used for Identifying Possible Event Types}} \\
     \midrule
        \multirow{5}{*}{Background} & In the context of open-domain event extraction, we aim to enhance the comprehensiveness of event annotations by supplementing existing annotations with additional potential event types. You are provided with a piece of text and its associated gold annotations, and the goal is to identify potential event types that may not have been annotated but could still exist within the text. This will help ensure that all relevant event types are captured in future analyses. \\
        ~ \\
        \multirow{4}{*}{Task Description} & You are tasked with analyzing the provided text and its gold annotations. Your goal is to identify any additional event types that might be present in the text but are not explicitly annotated. For each event type you identify, provide a fine-grained definition that captures the specific nuances of that event type. Before identifying event types, analyze the text to understand it. \\
        ~ \\
        \multirow{3}{*}{Output Format} & Provide your answer in the following Python dictionary format: \{"text\_analysis": "<your\_analysis>", "possible\_event\_types": [\{"event\_type": "<event\_type\_name>", "detailed\_definition": "<fine\_grained\_definition>"\}, ...]\}  \\
        ~ \\
        \multirow{11}{*}{Data} & Text: How is the world reacting to the war in Iraq? \\
        ~ & Gold trigger 0: war \# event type: Conflict:Attack \\ 
        ~ & Detailed definitions of gold event types: \\
        ~ & \quad Conflict:Attack: The "Conflict:Attack" event type involves an aggressive action or series of actions perpetrated by an Attacker with the intent to cause harm or exert control over a Target. The event may occur at a specific Place and can involve an Instrument, which is any tool or method used to carry out the attack. Additionally, an Agent may be present, indicating an entity responsible for orchestrating or executing the attack. Variations of this event type can range from physical confrontations, such as fights or battles, to assaults involving weapons or other means of inflicting harm, including psychological or chemical methods. The event is characterized by its hostile nature and the presence of a conflict between the involved entities. \\
     \bottomrule
    \end{tabularx}
    \end{threeparttable}
    \caption{Prompt used for identifying possible event types for each data instance in the benchmark. An example of data is illustrated.}
    \label{tab:identify_possible_event_types_prompt}
\end{table*}

\begin{table*}[h]
    \centering
    \small
    \begin{threeparttable}
    \begin{tabularx}{\textwidth}{cX}
     \toprule
        \multicolumn{2}{l}{\textbf{Prompt Used for Generating Supplementary Annotations}} \\
     \midrule
        \multirow{5}{*}{Background} & In the context of open-domain event extraction, we are supplementing existing annotations in text to ensure that all relevant event types are fully captured. You are provided with a piece of text, its associated gold annotations, and a list of possible event types with detailed definitions. Your task is to supplement the existing annotations by identifying additional event types that may be present in the text but not originally annotated. \\
        ~ \\
        \multirow{4}{*}{Task Description} & You are tasked with analyzing the provided text and its existing gold annotations. Based on the list of possible event types and their detailed descriptions, you need to supplement the annotations by identifying which of these additional event types are present in the text. Add these event types where appropriate, considering their triggers in the text. \\
        ~ \\
        \multirow{3}{*}{Output Format} & Please provide your answer in the following Python dictionary format: \{"text\_analysis": "<your\_analysis>", "supplemented\_annotations": [\{"event\_type": "<event\_type\_name>", "trigger\_word": "<word\_or\_phrase\_in\_text>"\}, ...]\}  \\
        ~ \\
        \multirow{11}{*}{Data} & Text: How is the world reacting to the war in Iraq? \\
        ~ & Gold trigger 0: war \# event type: Conflict:Attack \\ 
        ~ & Detailed definitions of gold event types: \\
        ~ & \quad Conflict:Attack: The "Conflict:Attack" event type involves an aggressive action ... (\textit{Omitted}) \\
        ~ & Possible event types with their detailed definitions: \\
        ~ & \quad Conflict:Demonstrate\_2: The event type refers to organized public gatherings ... (\textit{Omitted}) \\
        ~ & \quad contact.commandorder.correspondence\_1: The event type involves the act ... (\textit{Omitted}) \\
        ~ & \quad Reporting\_1: The Reporting event type encompasses instances where ...(\textit{Omitted}) \\
        ~ & \quad Earnings\_and\_losses\_1: The event type encompasses a wide range of scenarios ... (\textit{Omitted}) \\
        ~ & \quad Contact.Contact.Broadcast\_1: The Contact.Contact.Broadcast event type involves ... (\textit{Omitted}) \\
        ~ & \quad ... (\textit{Omitted other 16 event types along with their fine-grained definitions}) \\
     \bottomrule
    \end{tabularx}
    \end{threeparttable}
    \caption{Prompt used for generating supplementary annotations for every piece of data in the benchmark. An example of data is illustrated.}
    \label{tab:generate_supplementary_annotations_prompt}
\end{table*}

\begin{table*}[h]
    \centering
    \small
    \begin{threeparttable}
    \begin{tabularx}{\textwidth}{cX}
     \toprule
        \multicolumn{2}{l}{\textbf{Prompt Used for Merging Semantically Redundant Triggers after Nucleus Sampling}} \\
     \midrule
        \multirow{10}{*}{Background} & In the open-domain event extraction task, you are tasked with assisting in the process of merging duplicate event triggers within a dataset. The dataset has undergone multiple rounds of supplementation, resulting in 10 versions of triggers per data point. To ensure robustness and reliability, you have already performed a resampling process to select representative data points. However, during the event extraction task, subtle differences may occur in the mentions and event type definitions across repeated annotations. These differences need to be merged, reducing redundant information. Your job is to assist in merging similar event triggers that refer to the same event, based on their mention and event type definition. The triggers are divided into groups, and each group may contain multiple triggers that describe similar but not identical events. Your goal is to analyze and select the most appropriate trigger for a certain event, ensuring that only one trigger per event is retained.\\
        ~ \\
        \multirow{8}{*}{Task Description} & Given the original text and the resampled set of triggers with their corresponding fine-grained event type definitions, you are to perform the following steps: (1) Analyze the original text to understand the overall context and what events it discusses. (2) Analyze each group of triggers. Each group contains multiple triggers that potentially describe similar or related events, but they may not all refer to exactly the same thing. (3) For each group, assess the triggers based on their mentions and event type definitions. You need to determine which trigger best represents the event described and merge the similar triggers by selecting the most appropriate one. Output the merged triggers, ensuring that only the most relevant trigger for each event is retained. \\
        ~ \\
        \multirow{3}{*}{Output Format} &\{"Text Analysis": "<str: your analysis of the text>", "Group Analysis": ["<str: your analysis of the first group>", "<str: your analysis of the second group>", ...], "Merged Triggers": [\{"Trigger Span": "<str: the selected span in the text>", "Event Type": "<str: the event type corresponding to this trigger>"\}, ...]\}\\
        ~ \\
        \multirow{8}{*}{Data} & Text: The other option, and what the US seems to be trying, is to somehow fudge the election so Sistani does not win, that raise doubts about the outcome, claiming fraud, do anything so that the election causes to results. \\
        ~ & Group 0: \\
        ~ & \quad Trigger 0: claiming fraud \# event\_type: government.vote.violationspreventvote\_1 \\
        ~ & \quad Trigger 1: claiming fraud \# event\_type: contact.prevarication.broadcast\_1 \\
        ~ & \quad Trigger 2: fraud \# event\_type: government.vote.violationspreventvote\_1 \\
        ~ & Fine-grained Event Type Definitions: ... (\textit{Omitted fine-grained definitions of the above 2 event types}) \\
     \bottomrule
    \end{tabularx}
    \end{threeparttable}
    \caption{Prompt used for merging semantically redundant triggers for every piece of data after nucleus sampling. An example of data is illustrated.}
    \label{tab:merge_redundant_triggers_prompt}
\end{table*}

\subsection{Inference and Evaluation Prompts}
In our proposed SEOE evaluation framework, we leverage GPT-4o as the automatic evaluation agent to evaluate ODED abilities of open-source and closed-source LLMs. The inference prompt used for LLMs' inference is illustrated with an example in Table~\ref{tab:inference_prompt}. The evaluation prompt, which includes group information in Section~\ref{sec:groups}, is illustrated with an example in Table~\ref{tab:evaluate_prompt}.

\begin{table*}[h]
    \centering
    \small
    \begin{threeparttable}
    \begin{tabularx}{\textwidth}{cX}
     \toprule
        \multicolumn{2}{l}{\textbf{Prompt Used for LLMs' Inference}} \\
     \midrule
        \multirow{5}{*}{Background} & You are an event extractor designed to identify the presence of specific events in a given piece of text and to locate the corresponding event triggers in an open-domain setting. Open domain means there are no predefined constraints on the extraction process, and the text may originate from diverse sources and domains. This task requires comprehensive analysis and understanding of the text, as well as the ability to identify and classify events flexibly without reliance on a fixed ontology. \\
        ~ \\
        \multirow{7}{*}{Task Description} & In an open-domain setting, your task is to identify all possible triggers and their corresponding event types in the given text. A trigger is the key word or phrase in the text that most explicitly conveys the occurrence of an event. The process includes the following steps: (1) Analyze the text to understand its content and context. (2) Recognize potential event types that may be implied in the text. For each event type, generate a summarized name and provide a detailed description based on your cross-domain knowledge. (3) Extract triggers by identifying their spans in the text and linking them to their respective event types. Note that the text may contain zero, one, or multiple event triggers. \\
        ~ \\
        \multirow{5}{*}{Output Format} & Please provide your response in the following Python dictionary format: \{"Text Analysis": "<str: your analysis of the text>", "Event Type Generation": \{"<event\_type\_name\_1>": "<str: detailed\_description>", 
        "<event\_type\_name\_2>": "<str: detailed\_description>", ...\}, "Extraction Results": [\{"Trigger Span": "<str: a span in the text>", "Event Type": "<str: a specific event type from your generation>"\}, ...]\} \\
        ~ \\
        \multirow{1}{*}{Data} & Text: How is the world reacting to the war in Iraq? \\
     \bottomrule
    \end{tabularx}
    \end{threeparttable}
    \caption{Prompt used for large language models to do inference. An example of data is illustrated.}
    \label{tab:inference_prompt}
\end{table*}

\begin{table*}[h]
    \centering
    \small
    \begin{threeparttable}
    \begin{tabularx}{\textwidth}{cX}
     \toprule
        \multicolumn{2}{l}{\textbf{Prompt Used for Evaluating Open Domain Event Detection Models}} \\
     \midrule
        \multirow{9}{*}{Background} & You are an evaluation master designed to evaluate an open event detection model. In the open-domain event detection task, you aim to evaluate model performance at a semantic level. The evaluation target in open-domain event detection focuses on finding matching pairs of event triggers in the gold annotation set and the prediction set. A trigger refers to the key word or phrase that signals the occurrence of a specific event. Given an original text, a set of golden annotated triggers with event types, and a set of predicted triggers with event types, your goal is to match triggers with similar event types from both sets based on semantic equivalence. This evaluation does not require exact matches in trigger mention or event type definitions; rather, it focuses on identifying pairs of triggers that refer to the same event in the text and have semantically similar event type definitions. \\
        ~ \\
        \multirow{9}{*}{Task Description} & You are provided with an original text, a set of golden annotated triggers (with corresponding event types and definitions), a set of predicted triggers (also with event types and definitions), and a set of similar event type definitions used to assist in distinguishing event definitions. Your task is to: (1) Analyze the main event or events discussed in the original text. (2) Try to understand each golden trigger and each predicted trigger, along with their corresponding event type definitions. (3) Find several possible matching pairs of triggers that refer to the same event and have event type definitions that refer to the same event type. Then, analyze the event type definitions of two triggers' event types for each pair, and judge whether each pair can be regarded as a matching pair or not. (4) Provide the matching pairs of triggers' index as your final output. \\
        ~ \\
        \multirow{7}{*}{Output Format} & Please provide your response in the following Python dictionary format: \{"text\_analysis": "<str: your analysis>", "golden\_triggers\_understanding": ["<str: understanding for golden trigger 0>", "<str: understanding for golden trigger 1>", ...] , "predicted\_triggers\_understanding": ["<str :understanding for predicted trigger 0>", "<str: understanding for predicted trigger 1>", ...], "possible\_matching\_pairs": [
        \{"golden\_trigger": "<str: golden trigger mention \# its event type>", 
         "predicted\_trigger": "<str: predicted trigger mention \# its event type>", 
         "analysis\_and\_judgment": "<str: your analysis and judgment>"\}, 
        ...
        ], "final\_output": [\{"Gold\_trigger\_index": <int>, "Pred\_trigger\_index": <int>\}, ...]\}\\
        ~ \\
        \multirow{7}{*}{Data} & Text: How is the world reacting to the war in Iraq? \\
        ~ & Gold\_trigger\_0: war \# event\_type: Conflict:Attack\_1 \\
        ~ & Gold\_trigger\_1: reacting \# event\_type: Response\_1 \\
        ~ & Pred\_trigger\_0: reacting \# event\_type: Global Response \\
        ~ & Detailed definition of the above event types: ...(\textit{Omitted fine-grained definitions of the above event types}) \\
        ~ & Similar event of the above event types: ... (\textit{Omitted fine-grained definitions of event types within the same group as the above golden event types.}) \\
     \bottomrule
    \end{tabularx}
    \end{threeparttable}
    \caption{Prompt used for evaluating open domain event detection models. An example of data is illustrated.}
    \label{tab:evaluate_prompt}
\end{table*}


\end{document}